\documentclass{article}

\usepackage[preprint]{neurips_2020}

    \usepackage{neurips_2020}

\usepackage[utf8]{inputenc} 
\usepackage[T1]{fontenc}    
\usepackage{hyperref}       
\usepackage{url}            
\usepackage{booktabs}       
\usepackage{amsfonts}       
\usepackage{nicefrac}       
\usepackage{microtype}      
\usepackage{graphicx}
\usepackage{amsmath}

\title{Scope Loss for Imbalanced Classification and RL Exploration}

\author{%
  Hasham Burhani\thanks{hasham.burhani@rbccm.com, equal contribution}\\
  RBC CM\\
  \And
  Xiao Qi Shi\thanks{xiaoqi.shi@rbccm.com, equal contribution}\\
  RBC CM\\
  \And
  Jonathan Jaegerman\thanks{jonathan.jaegerman@gmail.com, equal contribution}\\
  University of Toronto\\
  \And
  Daniel Balicki\thanks{daniel.balicki@rbccm.com, equal contribution}\\
  RBC CM\\
}

\begin{document}

\maketitle

\begin{abstract}
  We demonstrate equivalence between the reinforcement learning problem and the supervised classification problem. We consequently equate the exploration exploitation trade-off in reinforcement learning to the dataset imbalance problem in supervised classification, and find similarities in how they are addressed. From our analysis of the aforementioned problems we derive a novel loss function for reinforcement learning and supervised classification. Scope Loss, our new loss function, adjusts gradients to prevent performance losses from over-exploitation and dataset imbalances, without the need for any tuning. We test Scope Loss against SOTA loss functions over a basket of benchmark reinforcement learning tasks and a skewed classification dataset, and show that Scope Loss outperforms other loss functions.
\end{abstract}

\section{Introduction}

\noindent Supervised learning has achieved state of the art performance in many domains over the past decade. Notably, supervised learning has recently achieved near-human performance in image classification  \cite{he2016deep, redmon2016you, wu2019detectron2, ramesh2022hierarchical, saharia2022photorealistic} and natural language processing \cite{2020t5, devlin2018bert, radford2018improving}. Meanwhile reinforcement learning (RL) has also made significant progress in gaming environments \cite{silver2016mastering, vinyals2019grandmaster, berner2019dota} and robotics \cite{MPABPMCE}. Although the problems of supervised classification and RL are dominated by similar deep learning solutions, they are treated as mutually exclusive subjects in academia and industry. We will first eliminate the barrier between them by framing the supervised classification problem as an RL problem. In other words, the set of all classification problems is a subset of all RL problems with several defining properties. Therefore, all of the terms describing a classification problem have one or more corresponding terms in an RL formulation of the same problem. As a lemma, we will draw the equivalence between the imbalanced dataset problem in supervised classification and the exploration exploitation trade-off in RL. \\

\noindent The imbalanced dataset problem is very common in many practical machine learning applications. The imbalanced dataset problem is commonly defined as the presence of a non-uniform distribution of classes in a supervised learning dataset. The dataset imbalance can cause performance losses as gradients from dominant classes can drown out gradients from other classes. In RL, it is important to balance exploitation with exploration because over-exploitation can cause the agent to get caught in sub-optimal policies. If an agent only exploits knowledge it already has, it will be unable to generate gradients towards new states and actions that could lead to better rewards. In other words, an exploitative agent generates an imbalanced dataset of the environment to learn from. \\

\noindent Over the past decade, many efforts have been devoted to coming up with solutions to both problems. Despite progress, the problems still persist. Notably, better loss functions have been invented to address the imbalanced dataset problem \cite{li2021autobalance, focal} and the exploration exploitation trade-off \cite{a3c, mnih2013playing}. We propose a novel loss function that inspires itself from solutions to both problems, and generalizes to work for both supervised classification and RL. 

\section{Imbalanced Dataset Problem vs Exploitation Problem in RL}

\noindent From here, we can note that the problem of dataset imbalance in supervised classification is equivalent to a lack of exploration in RL. An exploitative RL agent may generate an imbalanced dataset of the environment to learn from; limiting how optimal its policy can become. In classification, with an imbalanced dataset, the classification agent's policy might become too exploitative, repeatedly choosing actions that only work because of a slight statistical advantage in the environment. We can avoid this behaviour in both cases by promoting exploration, which would induce a robust understanding of the environment, from which the agent can form a more intelligent policy. As with any RL, exploration can be promoted with entropy loss functions and exploration rewards. \\

\begin{figure}
    \centering
    \includegraphics[width=0.75\textwidth]{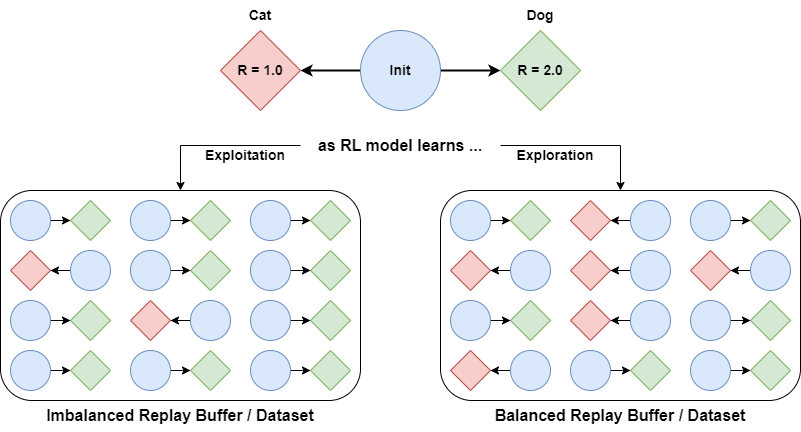}
    \caption{An illustration of how exploitation can generate imbalanced dataset.}
    \label{fig:self_reinforce_imbalance}
\end{figure}

\noindent With the following example, we demonstrate how an RL model can self-reinforce to generate an imbalanced dataset. Consider a binary path maze as illustrated in Figure \ref{fig:self_reinforce_imbalance}. The agent starts in the \(Init\) state and can decide to take the path to the left (\(Cat\)) or right (\(Dog\)). If the agent were to focus on exploration, there would be roughly equal number of Dog or Cat states in its replay buffer, i.e. the agent produces a balanced dataset of the environment. On the other hand if the agent were to focus on exploitation, knowing it can receive more reward from the \(Dog\) state, the agent would quickly converge to mostly taking the right path. Thus, the replay buffer would be filled with more \(Dog\) states than \(Cat\) states, i.e. the agent produces an imbalanced dataset of the environment. 

\noindent Despite equivalence, results from research regarding each problem are not shared and both problems are solved quite differently. Notably, loss function adaptations are completely inconsistent.

\section{Background}

\noindent\textbf{Imbalanced Dataset Classification:} The problem of imbalanced datasets in classification is most commonly addressed with hard negative mining \cite{nm1236, DBLP:journals/corr/ShrivastavaGG16}. With hard negative mining, difficult or underrepresented training examples are resampled or weighted stronger gradients. New loss functions have also been created as a robust method to automatically adjust the weights of training sample gradients relative to how easily the model classifies them. Modified Huber Loss \cite{ESLII, https://doi.org/10.48550/arxiv.2108.12627} combines properties of both MSE Loss and MAE Loss to down-weigh the gradients of outliers and well-classified examples based on the magnitude of their error. Designed specifically for the imbalanced object detection problem, Focal Loss \cite{https://doi.org/10.48550/arxiv.1708.02002} down-weighs gradients of well-detected training examples based on the magnitude of the probability of the detection.\\

\noindent\textbf{RL Exploration:} RL exploration can be promoted via careful design of intrinsic reward functions \cite{DBLP:journals/corr/abs-1908-06976}. Using an environment model,  \cite{https://doi.org/10.48550/arxiv.1611.07507} promotes exploration with intrinsic rewards assigned to policies that can reach more states. DORA \cite{DBLP:journals/corr/abs-1804-04012} learns a secondary MDP in parallel; this MDP carries rewards that surrogate the knowledge value of state-actions. NGU \cite{https://doi.org/10.48550/arxiv.2002.06038} uses episodic memory to reward trajectories that don't revisit states. An entropy term in RL loss functions has recently emerged as a robust method for promoting exploration with minimal tuning \cite{MPABPMCE}. \\

\noindent\textbf{RL for Classification and Regression:} With a significant increase in the performance of RL over the last decade, RL has recently been applied to the mainstream problems of classification and regression. By repeatedly training neural networks, framing model design and hyperparameter selection as RL actions, and using validation accuracy as a reward function, RL has seen success addressing the problems of neural architecture search \cite{https://doi.org/10.48550/arxiv.1611.01578}, hyperparameter optimization \cite{https://doi.org/10.48550/arxiv.1906.11527} and transfer learning \cite{mahmud2022rltune}. With a considerable loss in training efficiency, RL has also been used to directly train neural networks for high generalization performance in complex classification tasks \cite{DBLP:journals/corr/abs-1901-01379, RLAfsCP, Zhang_Huang_Zhao_2018, Zhao2017DeepRL, DBLP:journals/corr/abs-1912-09595}. In these examples, the supervised classification problem is framed as an RL problem by treating an attempted classification as an action, and generating a reward function from data labels.

\subsection{Imbalanced Datasets in Supervised Classification}

The imbalanced dataset problem is present in supervised classification environments with dataset distribution imbalances. Although this definition is widely used and captures most examples, it does fail under two scenarios. First, a model will have no problem classifying a trivial dataset regardless of whether the dataset distribution is skewed. Second, a non-skewed dataset with some classes that are much harder to classify than others will cause a similar performance loss. Consider, for example, a dataset with an equal number of class A and class B labelled images, but the quality of the class A images is much worse than the quality of the class B images. We propose to define the imbalanced dataset problem generally as follows: given a model and a dataset, the imbalanced dataset problem is present when classes are disparately difficult to classify, resulting in the model performing sub-optimally on some classes. \\

\noindent With a dataset imbalance, a classification agent can learn to perform relatively well by exploiting the imbalance and predicting easier classes, rather than learning a smarter optimal policy that maximizes predictions over each class. In supervised classification, the common solution is to artificially balance the dataset with hard negative mining, data augmentation or adaptive loss functions that balance gradients.

\subsubsection{Focal Loss}
Cross Entropy Loss is the original loss function used for supervised multi-class classification.
\begin{equation} \label{eq:ce}
\text{Cross Entropy Loss (Supervised)} = -log(p_{i})
\end{equation}
Where \(p_{i}\) is the probability of predicting the target class. Focal loss is an adaptive loss function that balances gradients by reducing the magnitude of gradients from well-classified examples. Focal Loss applies a scaling term to Cross Entropy Loss that decays to zero as the probability of the target class increases to one.
\begin{equation} \label{eq:fl}
\text{Focal Loss (Supervised)} = -(1-p_{i})^{\gamma}log(p_{i})
\end{equation}
Where \(\gamma \geq 0\) is a focusing parameter, which controls the curvature of the scaling term as \(p_{i}\) varies from zero to one.

\subsection{Exploration Exploitation in RL}

The exploration exploitation trade-off is a difficult problem in RL. RL agents learn an optimal behavior from environments by experience; at the same time, an agent's behaviour determines what it will experience from the environment. To maximize the quality of learning from an environment, the RL agent must explore the environment and discover how it can receive the most reward. To maximize returns from an environment, the RL agent must instead exploit what it already knows to receive reward. In RL algorithms, we generally find an agent that explores near the start of its training and exploits more as it understands the environment. This agent, however, often explores too much and never fully converges to an efficient exploitative policy, or the agent exploits too much and settles at a sub-optimal policy. Vanilla RL algorithms are mostly exploitative and have trouble performing well in environments with sparse reward functions that require a lot of exploration.
 
\subsubsection{Policy and Entropy Loss}
Policy Loss is one of the original loss functions used for Actor Critic RL algorithms.

\begin{equation} \label{eq:pl}
\text{Policy Loss (Actor Critic)} = -Alog(p_{i})
\end{equation}
Where \(A\) is the advantage term in the Actor Critic setup, and \(p_{i}\) is the probability of selecting the chosen action. One common method of encouraging exploration in Actor Critic algorithms is to introduce an entropy term into the loss function. The (negative) entropy term is minimized when all action probabilities are equal. Whereas the policy term maximizes exploitation of known environmental advantages by increasing the probability of actions that result in a high advantage, the entropy term promotes exploration by blindly increasing the probability of all actions.
\begin{equation} \label{eq:pel}
\begin{split}
\text{Policy and Entropy Loss (Actor Critic)} & = -Alog(p_{i}) + \alpha \sum_{j} p_{j}log(p_{j}) \\
    & = -Alog(p_{i}) + \alpha p_{i}log(p_{i}) + \alpha \sum_{j,j \neq i} p_{j}log(p_{j}) \\
    & = -(A-\alpha p_{i})log(p_{i}) + \alpha \sum_{j,j \neq i} p_{j}log(p_{j}) 
\end{split}
\end{equation}
Where \(\alpha \geq 0\) adjusts the magnitude of the entropy term, which increasingly encourages exploration as \(\alpha\) increases. Policy and Entropy Loss contains bias in its equation from the \(\alpha \sum_{j,j \neq i} p_{j}log(p_{j})\) term. Namely, if \(\alpha>0\), the optimal action's probability will never converge to \(1\), as the entropy term will not be minimized.

\subsection{Supervised Classification as RL}

RL environments are usually modeled as a Markov Decision Process (MDP), and most RL algorithms are designed around MDP concepts. The classification task can easily be modeled as an MDP. The unlabelled data comprises of a Markov state \(s\) from which the agent will choose to classify the data as any of the available classes. The act of classifying is equivalent to a Markov action \(a\) selected from a discrete action space, and a reward \(r\) is provided based on the correctness of the classification action. Given training example labels, the reward function can be known before having taken an action. \\

\noindent The classification environment can be framed as a one-step episodic MDP, in which each new image is the start of a new episode, with a single action to be immediately followed by a reward and a terminal state \(S = (s_{0}, s_{T})\), \(A = (a_{0})\), \(G = R = (r_{0})\). The environment can also otherwise be framed as a continuous MDP, with random state transitions, for which a non-zero discount factor is redundant \(\gamma = 0\), \(S = (s_{0}, s_{1}, ...)\), \(A = (a_{0}, a{1}, ...)\), \(G = R = (r_{0}, r{1}, ...)\). The classification environment can therefore be generalized as the subset of MDPs with the following three key properties:
\begin{enumerate}
  \item A discrete action space.
  \item A known reward function.
  \item Episodic and all states transition to the terminal state OR continuous and all state transitions are random.
\end{enumerate}
With this MDP model, the classification environment can be approached as an RL problem and solved using standard RL methods. Furthermore, by setting \(A=1\) and \(i=\text{index of the correct class}\), we can apply Policy and Entropy Loss to classification. 
\begin{equation} \label{eq:pels}
\text{Policy and Entropy Loss (Supervised)} = -log(p_{i}) + \alpha \sum_{j} p_{j}log(p_{j})
\end{equation}
In fact, if we also set \(\alpha=0\) (i.e. remove the entropy term), the Policy Loss will be exactly equivalent to Cross Entropy Loss \ref{eq:ce}. Existing supervised classification training methods can be interpreted as sample efficiency and training stability tricks that exploit classification's unique defining properties. Conversely, Focal Loss can be extended to Actor Critic RL algorithms by injecting an advantage term \(A\) into the scaling term, and an additional hyperparameter \(\alpha\) to scale \(p_i\) w.r.t. \(A\).
\begin{equation} \label{eq:flrl}
\text{Focal Loss (Actor Critic)} = -(A-\alpha p_{i})^{\gamma}log(p_{i})
\end{equation}

\section{Method}

\subsection{Scope Loss}
We inspire ourselves from Focal Loss and Policy and Entropy Loss and present Scope Loss. Like Focal Loss, Scope Loss prevents performance losses from class imbalances by balancing class gradients relative to certainty. Equivalently, like Policy and Entropy Loss, Scope Loss promotes exploration by avoiding high probability policies.
\begin{equation} \label{eq:sl}
\text{Scope Loss (Actor Critic)} = -(A-\alpha p_{i})log(p_{i})
\end{equation}
Where \((A-\alpha p_{i})\) is the scope scaling factor, and \(\alpha\) is the scope hyperparameter that scales \(p_{i}\) relative to the magnitude of \(A\) (which varies from environment to environment). Where Policy and Entropy Loss promotes exploration by blindly incorporating an arbitrary amount of exploration into the agent, Scope Loss indirectly promotes exploration by setting an upper limit on how quickly an agent can become exploitative. When \(A>0\) a greater certainty \(p_i\) will decrease the gradient, thereby preventing exploitation. Note than Scope Loss can be derived by removing the bias term in \ref{eq:pel}, which allows the agent to theoretically reach the optimal policy. We can also derive Scope Loss by setting \(\gamma=1\) in \ref{eq:fl} (removing the additional hyperparameter). For supervised classification we can set \(A=1\) and \(\alpha=1\). 
\begin{equation} \label{eq:sls}
\text{Scope Loss (Supervised)} = -(1-p_{i})log(p_{i})
\end{equation}
Determining the value of \(\alpha\) for RL is more tricky, because the advantage term A distribution is not constant between environments or throughout training. \(\alpha\) should be small enough that the scope term is still dominated by the advantage term. Instead of manually scheduling and tuning \(\alpha\) for each environment, we can instead fix \(\alpha\) and employ z-score batch normalization of advantages. With this approach we can find a single optimal value for \(\alpha\) for any environment and throughout all stages of training.
\begin{equation} \label{eq:anorm}
A := \frac{A - \mu_{A}}{\sigma_{A}}
\end{equation}

\subsection{Scope Loss for PPO}

We defined Scope Loss for Actor Critic RL algorithms and supervised classification. PPO RL algorithms have proven very effective in distributed settings so would like to extend Scope Loss to PPO algorithms. We can derive a PPO equivalent following a similar procedure. We start from vanilla Policy Loss for PPO.
\begin{equation} \label{eq:policyppo}
\text{Policy Loss (PPO)} = -\min(\frac{p_i}{p^k_i}A^k, \ g(\epsilon, A^k)), \text{ where } g(\epsilon, A) = \begin{cases}
        A(1+\epsilon) & \text{if } A \ge 0\\
        A(1-\epsilon) & \text{if } A < 0
    \end{cases}
\end{equation}
Where \(i\) indicates the index of the chosen action, \(k\) indicates the index of the worker model that is using its experience to update the main model, and \(\epsilon\) is the PPO clipping sensitivity hyperparameter that determines how different the worker and main policies must be for the loss gradient to be truncated. We can now add an entropy term for a PPO Policy and Entropy Loss.
\begin{equation} \label{eq:policyentropyppo}
\begin{split}
\text{Policy and Entropy Loss (PPO)} & = -\min(\frac{p_i}{p^k_i}A^k, \ g(\epsilon, A)^k) + \alpha \sum_{j} p_{j}log(p_{j}) \\
    & = -\min(\frac{p_i}{p^k_i}A^k, \ g(\epsilon, A^k)) + \alpha p_{i}log(p_{i}) + \alpha \sum_{j,j \neq i} p_{j}log(p_{j})
\end{split}
\end{equation}
Here, due to the \(\min\) term, we cannot factorize to reproduce the scope factor. By letting \(\alpha = 0\) we can arrive at Scope Loss for PPO.

\begin{equation} \label{eq:scopeppo}
\text{Scope Loss (PPO)} = -\min(\frac{p_i}{p^k_i}A^k, \ g(\epsilon, A^k)) + \alpha p_{i}log(p_{i})
\end{equation}

\noindent When \(\frac{p_i}{p^k_i}A^k\) results as the minimum, we can reformulate to isolate the PPO version of the scope scaling factor.

\begin{equation} \label{eq:scopepposimp}
\begin{split}
& = -\frac{p_i}{p^k_i}A + \alpha p_{i}log(p_{i}) \\ & = -\frac{p_i}{p^k_i}(A - \alpha p_{i}^klog(p_{i}))
\end{split}
\end{equation}

\noindent Where \((A - \alpha p_{i}^klog(p_{i}))\) is the scope scaling factor. In this case, the scope scaling factor incorporates the cross entropy of the main model policy w.r.t. the worker model policy. The scope scaling factor will balance gradients such that:
\begin{enumerate}
    \item When main model certainty \(p_{i}\) is large and \(A>0\), the gradient is decreased to prevent over-exploitation.
    \item When worker model certainty \(p_{i}^k\) is smaller than main model certainty \(p_{i}\) and \(A>0\), the gradient is decreased to limit the distance between the models.
    \item When worker model certainty \(p_{i}^k\) is greater than main model certainty \(p_{i}\) and \(A<0\), the gradient is decreased to limit distance between the models.
\end{enumerate}
This gradient balancing incorporates ideas from both PPO Loss design and Scope loss design. \\

\noindent We also extend Focal Loss to the PPO setting, for a fair comparison in our experiments, by fixing \(\gamma=2\) (one of the recommended values in the original paper), expanding the square in \ref{eq:flrl}, and replacing any resulting Actor Critic policy loss terms \ref{eq:pl} with PPO Policy Loss \ref{eq:policyppo}.

\begin{equation} \label{eq:focalppo}
\text{Focal Loss (PPO)} = -\min(\frac{p_i}{p^k_i}A^k, \ g(\epsilon, A^k))A + 2\alpha\min(\frac{p_i}{p^k_i}A^k, \ g(\epsilon, A^k))p_{i} - \alpha^2p_i^2log(p_i)
\end{equation}

\section{Results}

We first compare Scope Loss with Focal Loss for RL and supervised classification, and use Policy Loss and Policy and Entropy Loss as baselines. We demonstrate that Scope Loss generally performs best in both settings with no tuning, whereas FL fails to learn a meaningful policy in RL. We ran all of the experiments on a single V100 NVIDIA GPU.

\subsection{RL}

The selected environments include a basket of Atari benchmarks; where Asterix-V0, Seaquest-v0, Venture-v0, Gravitar-v0 and MontezumaRevenge-v0 were selected to test for sufficient exploration. Procgen StarPilot \cite{DBLP:journals/corr/abs-1912-01588} is also included to test for generalization performance. We train the Impala architecture \cite{DBLP:journals/corr/abs-1802-01561} using PPO with the hyperparameters outlined in \ref{table:rlhyperparameter}. We run each trial on the same environment seed using the PPO loss functions introduced in the previous section: Policy Loss \ref{eq:policyppo}, Policy and Entropy Loss \ref{eq:policyentropyppo}, Scope Loss \ref{eq:scopeppo}, and Focal Loss \ref{eq:focalppo}. \\

\begin{figure}
    \includegraphics[width=0.23\textwidth]{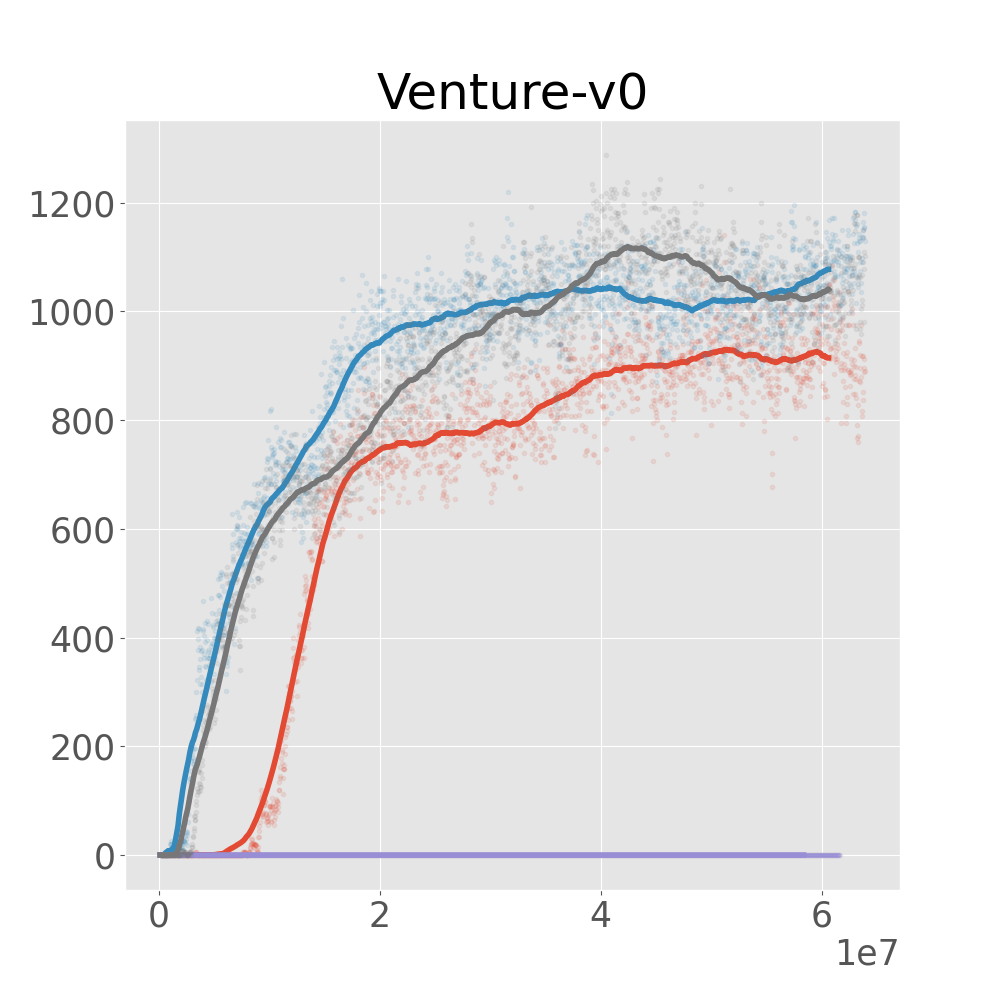}
    \includegraphics[width=0.23\textwidth]{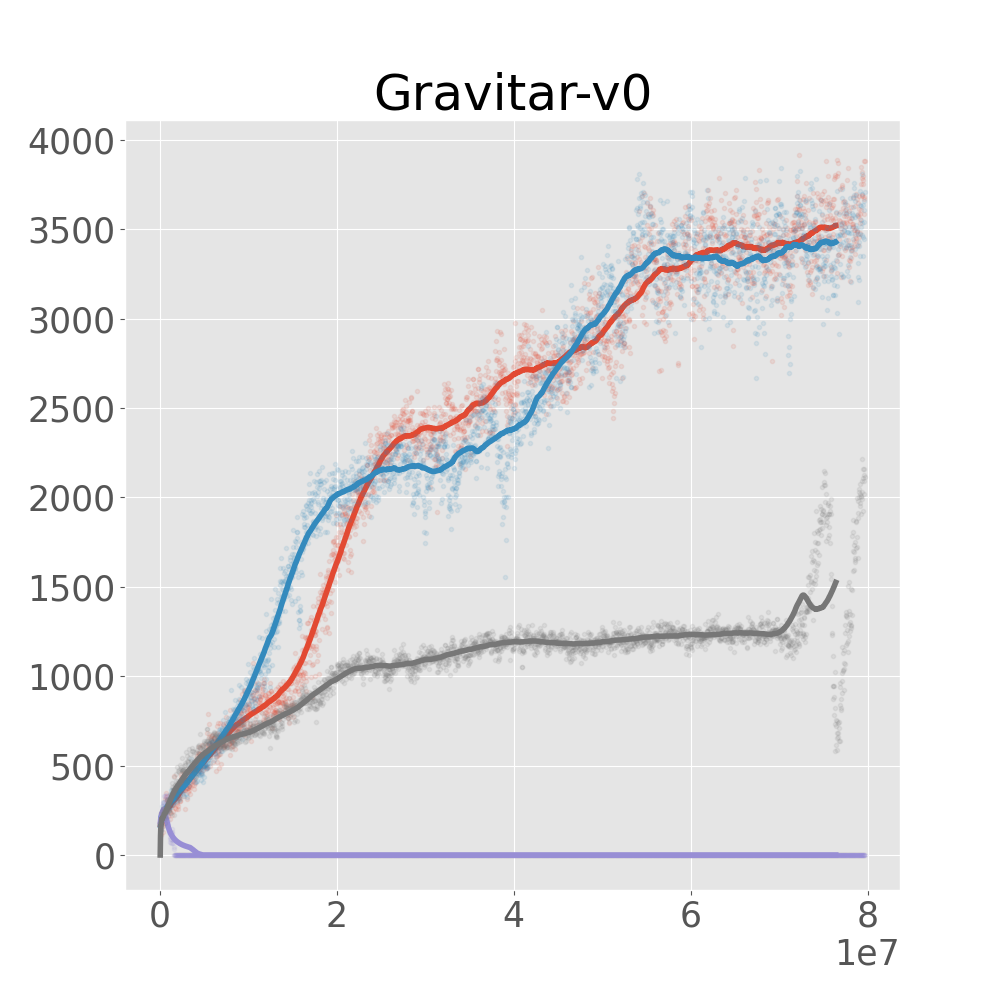}
    \includegraphics[width=0.23\textwidth]{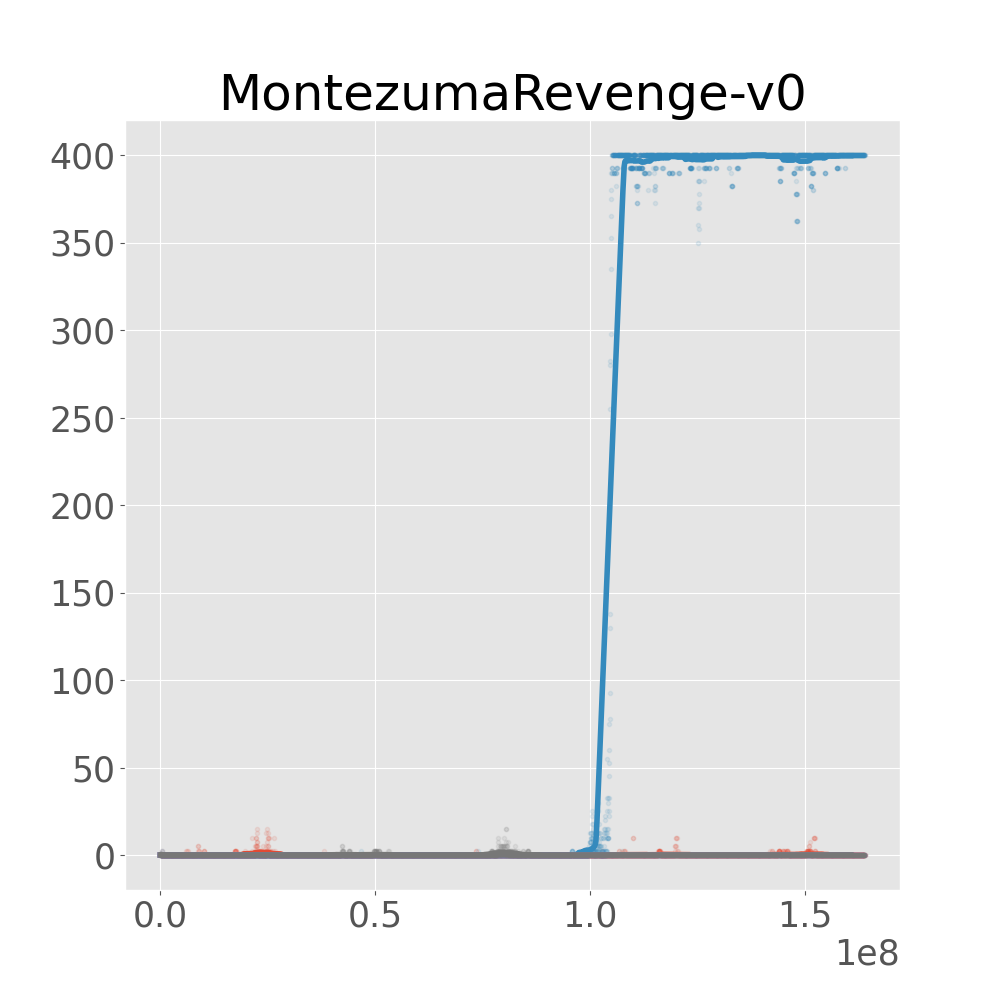}
    \includegraphics[width=0.23\textwidth]{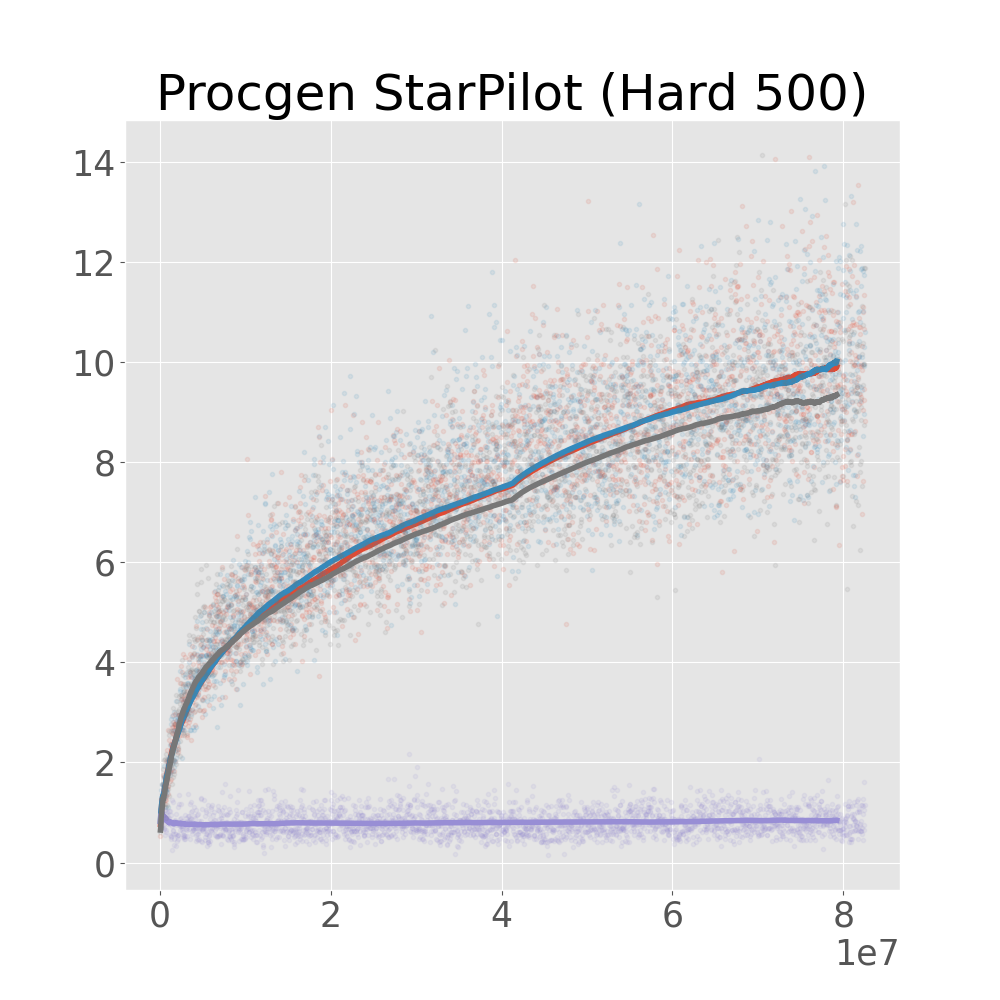}
    \includegraphics[width=0.9\textwidth]{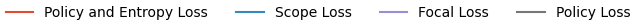}
    \centering\caption{RL Loss Comparison Training Curves}
    \label{fig:RL}
\end{figure}

\noindent We also estimate advantages with the Generalized Advantage Estimation (GAE) formulation \cite{https://doi.org/10.48550/arxiv.1506.02438} and apply advantage batch normalization \ref{eq:anorm}. We can see in Figure \ref{fig:RL} that Focal Loss consistently under-performs, and fails to generalize to RL. We can also see that Scope Loss generally outperforms the other loss functions across a variety of environments, usually taking off the fastest and finishing and generating one of the top two final policies. Scope Loss, like Policy and Entropy Loss, finds better policies via exploration, but Scope Loss also consistently finishes with a slight advantage. We deduce this final advantage to be a result of the removal of the additional bias in the Policy and Entropy Loss.
\\
\\
\begin{figure}
    \centering
    \includegraphics[width=0.23\textwidth]{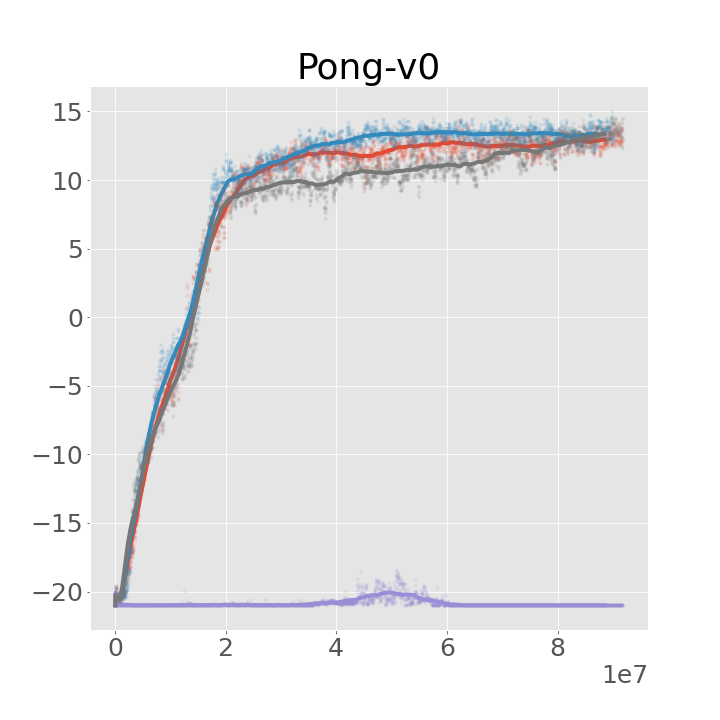}
    \includegraphics[width=0.23\textwidth]{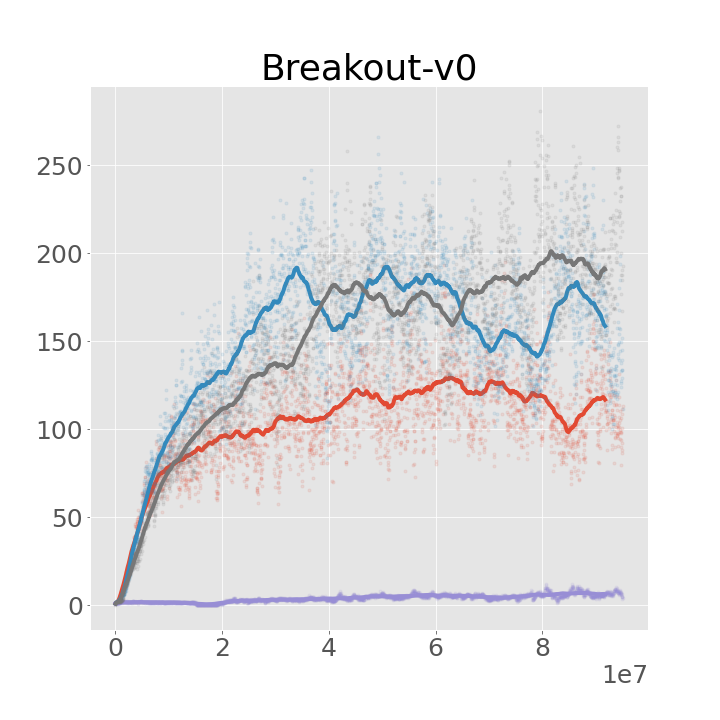}
    \includegraphics[width=0.23\textwidth]{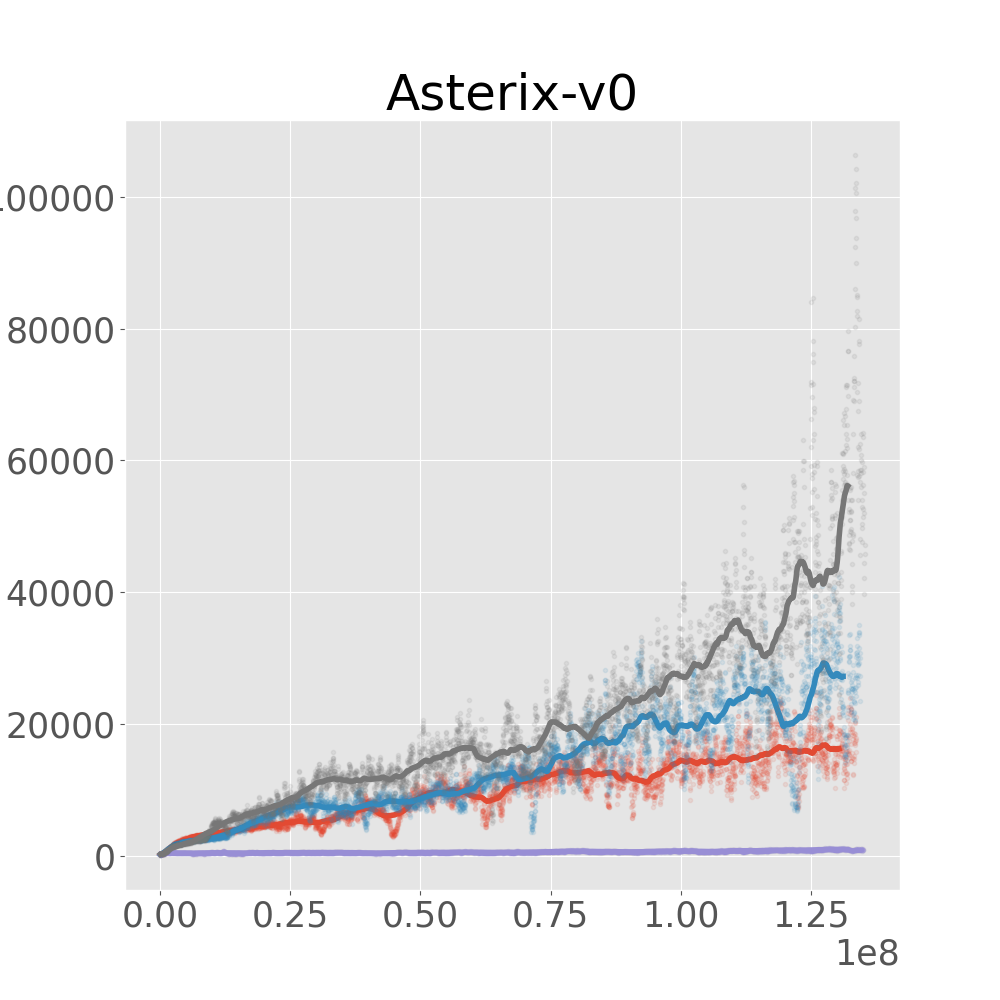}
    \includegraphics[width=0.23\textwidth]{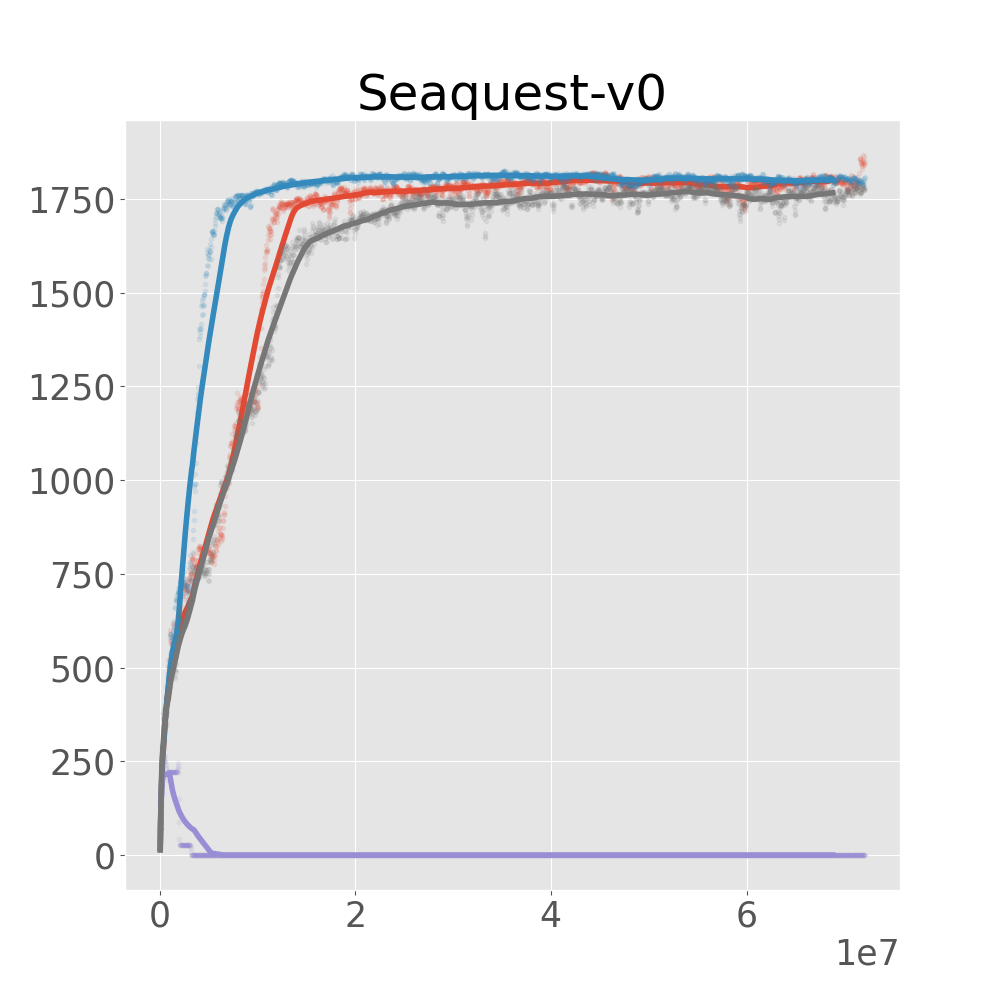}
\end{figure}

\subsection{Supervised Classification}

We chose the Caltech-256 \cite{griffin_holub_perona_2022} dataset as our benchmark for supervised classification. Caltech-256 contains 30607 images, of 257 classes with strong class imbalances. Notably, the classes have a minimum count of 80 examples and a maximum of 827. We train a ResNet101 model from scratch using an Adam optimizer with the hyperparameters outlined in \ref{table:schyperparameter}. We run each trial on the same dataset split using the supervised classification loss functions introduced in the previous sections: Cross Entropy Loss (CEL) \ref{eq:ce}, Focal Loss (FL) \ref{eq:fl}, Policy and Entropy Loss (PEL) \ref{eq:pels}, and Scope Loss (SL) \ref{eq:sls}. We train until convergence and keep the model checkpoint with the highest validation accuracy. We then evaluate each trial's best model over a test set and compute their precision over each class. \\

\noindent We can see in the table below that Scope Loss achieves the greatest mean precision, with a 20\% relative improvement compared to the other loss functions. We can also see that Policy and Entropy Loss achieves the lowest standard deviation of precision, with a 17\% relative improvement compared to the other loss functions. Where Policy and Entropy Loss attempts to maximize all class precisions equally, by forcing the classification agent to consider predicting unlikely classes; Scope Loss and Focal Loss attempt to maximize all class precisions independently without being blinded by class imbalances, by weighing gradients dynamically, and avoiding exploitation of the easiest classes. This distinction enables Scope Loss to promote exploration without risking losses in performance.

\begin{table}[h]
\centering
\def\arraystretch{1.5}
\begin{tabular}{ |c||c|c|c|c|}
\hline
    \textbf{Precision Metric} & \textbf{PEL} & \textbf{SL} & \textbf{FL} & \textbf{CEL}  \\
\hline
    Mean  & \(0.2303\) & \(0.2814\) & \(0.2320\) & \(0.2383\) \\
    Standard Deviation & \(0.2116\) & \(0.2436\) & \(0.2360\) & \(0.2471\) \\
\hline
\end{tabular}
\caption{Supervised Classification Loss Comparison Precision Metrics}
\label{table:superclassexper1}
\end{table}

We repeat the experiment with a suite of image augmentation techniques and manual SGD learning rate tuning and scheduling. We employ a cosine decay scheduler with hyperparameters outlined in \ref{table:schyperparameter2}. We train for 100 of epochs, over which the cosine scheduler decays the learning rate to 0. We run a separate trial for the same loss functions and test the best checkpoint models on a test set.\\

\noindent We can see the same patterns emerge in the table below. Scope Loss results in the greatest mean precision and Policy and Entropy Loss results in the lowest standard deviation of precision. In this experiment, however, the differences in precision metrics between loss trials is much less significant. This decrease in the effect of our exploratory loss functions is expected, as the data augmentation effectively reduces the dataset imbalance. The augmentation techniques remove any "obvious" features that the classification agent could use to exploit otherwise easier classes.

\begin{table}[h]
\centering
\def\arraystretch{1.5}
\begin{tabular}{ |c||c|c|c|c|}
\hline
    \textbf{Precision Metric} & \textbf{PEL} & \textbf{SL} & \textbf{FL} & \textbf{CEL}  \\
\hline
    Mean  & \(0.5823\) & \(0.5956\) & \(0.5723\) & \(0.5929\) \\
    Standard Deviation & \(0.2055\) & \(0.2195\) & \(0.2227\) & \(0.2196\) \\ 
\hline
\end{tabular}
\caption{Augmented Supervised Classification Loss Comparison Precision Metrics}
\label{table:superclassexper2}
\end{table}

\section{Conclusion}
We generalized the problem of supervised classification dataset imbalances to the reinforcement learning (RL) exploration exploitation trade-off, and introduced Scope Loss as a generalization of the loss functions used to solve these two problems. Scope Loss indirectly promotes exploration by preventing over-exploitation and can easily be applied to both supervised classification and RL. Scope Loss inspires itself from Policy and Entropy Loss and Focal Loss, but it only carries a single hyperparameter that requires no tuning, and it removes the theoretical bias in Policy and Entropy Loss. We then demonstrated that Scope Loss is a robust loss function that outperforms Focal Loss, Cross Entropy and Policy Loss with and without Entropy in both RL and supervised classification.

\newpage

\bibliography{bib}

\newpage

\section{Appendix}
\subsection{RL Hyperparameters}

\begin{table}[h]
\centering
\begin{tabular}{ |c||c|c|c|c|c|c|c|}
\hline
       Environment & $lr$ & ${\alpha}_{value}$ & ${\alpha}_{entropy}$ & ${\alpha}_{scope}$ & ${\alpha}_{focal}$ & $\gamma$ & $\lambda$\\
\hline
    Pong-v0  & \(10^{-4}\) & 0.5 & 0.01 & 0.01 & 2 & 0.999 & 0.95 \\
    Breakout-v0 & \(10^{-4}\) & 0.5 & 0.01 & 0.01 & 2 & 0.999 & 0.95 \\
    Asterix-v0 & \(10^{-4}\) & 0.5 & 0.01 & 0.01 & 2 & 0.999 & 0.95 \\
    Seaquest-v0 & \(10^{-4}\) & 0.5 & 0.01 & 0.01 & 2 & 0.999 & 0.95 \\
    Venture-v0 & \(10^{-4}\) & 0.5 & 0.01 & 0.01 & 2 & 0.999 & 0.95 \\
    Gravitar-v0 & \(10^{-4}\) & 0.5 & 0.01 & 0.01 & 2 & 0.999 & 0.95 \\
    MontezumaRevenge-v0 & \(10^{-4}\) & 0.5 & 0.01 & 0.01 & 2 & 0.999 & 0.95 \\
    Procgen StarPilot (Hard 500) & \(10^{-4}\) & 0.5 & 0.01 & 0.01 & 2 & 0.999 & 0.95 \\
    Caltech 256 Classification & \(10^{-3}\) & 0.5 & 0.05 & NA & NA & 0 & 0\\
\hline
\end{tabular}
\caption{Hyperparameters used for all RL experiments.}
\label{table:rlhyperparameter}
\end{table}

\subsection{Supervised Classification Hyperparameters}
\begin{table}[h]
\centering
\begin{tabular}{ |c|c|c|c|c|c|c|}
\hline
    $lr$ & $\alpha_{scope}$ & $\alpha_{entropy}$ & $\alpha_{focal}$\\
\hline
    \(10^{-3}\) & 1 & \(8*10^{-6}\) & 2 \\
\hline
\end{tabular}
\caption{Hyperparameters used in the first Supervised Classification experiment.}
\label{table:schyperparameter}
\end{table}
\begin{table}[h]
\centering
\begin{tabular}{ |c|c|c|c|c|c|c|c|c|c|c|}
\hline
    $lr$ & $epochs$ & $\alpha_{scope}$ & $\alpha_{entropy}$ & $\alpha_{focal}$ & $color$\\
\hline
    \(10^{-2}\) & 100 & 1 & \(8*10^{-6}\) & 2 & \(\mathcal{N}(1,0.3)\)\\
\hline
    contrast & brightness & sharpness & crop & flip &  \\
\hline
    $\mathcal{N}(1,0.1)$ & $\mathcal{N}(1,0.1)$ & $\mathcal{N}(1,0.3)$ & $\mathcal{U}$ & $Bernoulli(0.5)$ & \\
\hline
\end{tabular}
\caption{Hyperparameters used in the augmented Supervised Classification experiment.}
\label{table:schyperparameter2}
\end{table}

\end{document}